%% file: emnlp2021.tex
\pdfoutput=1

\documentclass[11pt]{article}

\usepackage[]{emnlp2021}
\usepackage{algorithm}
\usepackage{algorithmic}
\usepackage{times}
\usepackage{latexsym}

\usepackage[T1]{fontenc}

\usepackage[utf8]{inputenc}

\usepackage{microtype}

\usepackage{graphicx}
\usepackage{amsmath}
\usepackage{multirow}

\usepackage{comment} 
\newcommand{\cut}[1]{}

\newcommand{\at}[1]{\textcolor{magenta}{\textbf{#1}}}

\newcounter{tbsnr}
\newenvironment{tbs}
{\addtocounter{tbsnr}{1}\par\bigskip\noindent\fbox{\thetbsnr}
\hspace*{\fill}\begin{minipage}{7cm}\tt}
{\end{minipage}\hspace*{\fill}\bigskip}

\title{Looking for Confirmations:\\ An Effective and Human-Like Visual Dialogue Strategy} 

\author{Alberto Testoni \\
	DISI - University of Trento \\
	Trento - Italy \\
	\texttt{alberto.testoni@unitn.it} \\\And
	Raffaella Bernardi \\
	CIMeC and DISI - University of Trento\\
	Rovereto (TN) - Italy \\
	\texttt{raffaella.bernardi@unitn.it} \\}

\begin{document}
\maketitle
\begin{abstract}

Generating goal-oriented questions in Visual Dialogue tasks is a challenging and long-standing problem. State-Of-The-Art systems are shown to generate questions that, although grammatically correct, often lack an effective strategy and sound unnatural to humans. Inspired by the cognitive literature on information search and cross-situational word learning, we design \textit{Confirm-it}, a model based on a beam search re-ranking algorithm that guides an effective goal-oriented strategy by asking questions that confirm the model's conjecture about the referent. We take the GuessWhat?!\@ game as a case-study. We show that dialogues generated by \textit{Confirm-it} are more natural and effective than beam search decoding without re-ranking.


\end{abstract}

\input{introduction}
\input{related}

\input{task_data}
\input{model_experiments}

\input{results}

\input{conclusion}

\section*{Acknowledgements}
The authors kindly acknowledge the support of NVIDIA Corporation with the donation of the GPUs used in our research. We would like to thank Albert Gatt, Stella Frank, Claudio Greco, and Michele Cafagna for their suggestions and comments. Finally, we thank the anonymous reviewers for the insightful feedback.

\bibliography{anthology,custom,raffa}
\bibliographystyle{acl_natbib}

\appendix
\section{Supplementary Material}

\begin{figure}[b]\centering
	\includegraphics[width=1\linewidth]{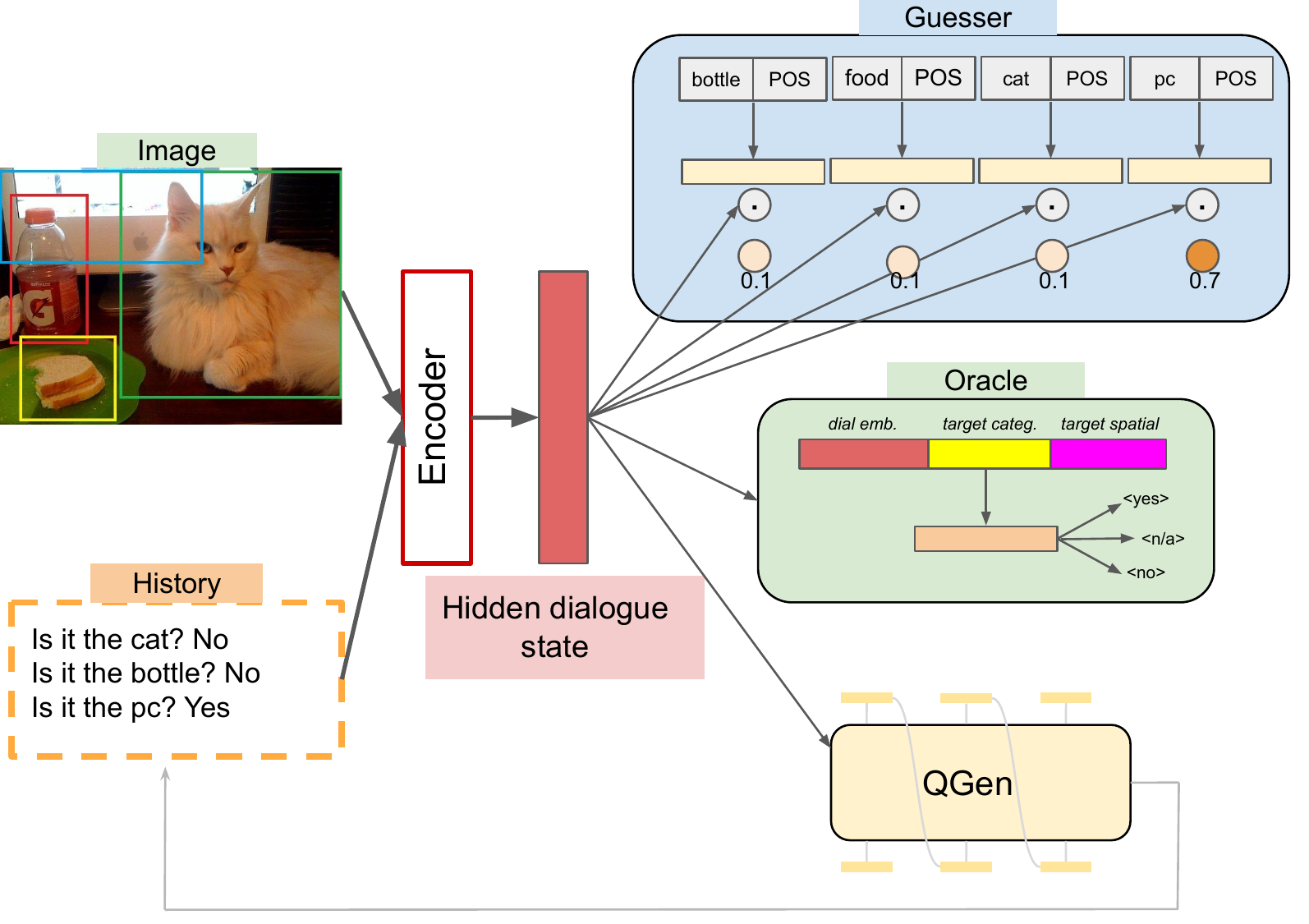}
	\caption{Model architecture of \textit{Confirm-it}.}
	\label{fig:gdse_v3}
\end{figure}

\begin{figure}[]\centering
	\includegraphics[width=1\linewidth]{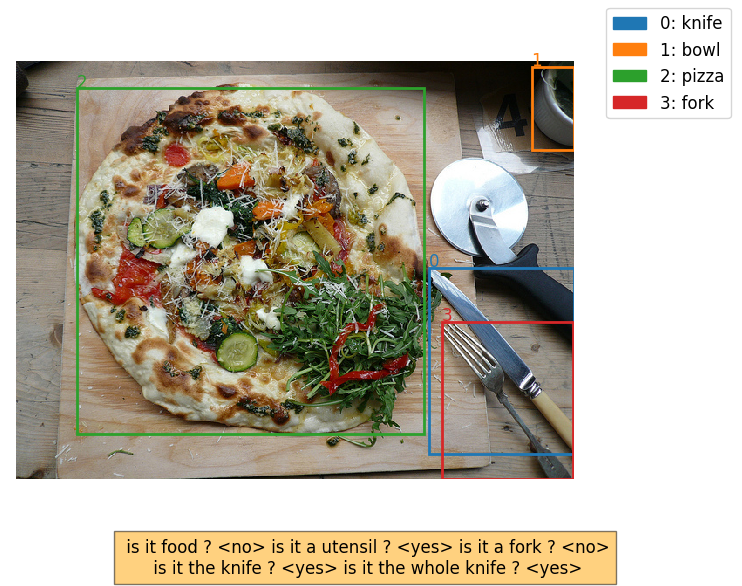}
	\caption{Annotation carried out by human participants.}
	\label{fig:annotation_human}
\end{figure}

\begin{figure}[b]\centering
	\includegraphics[width=1\linewidth]{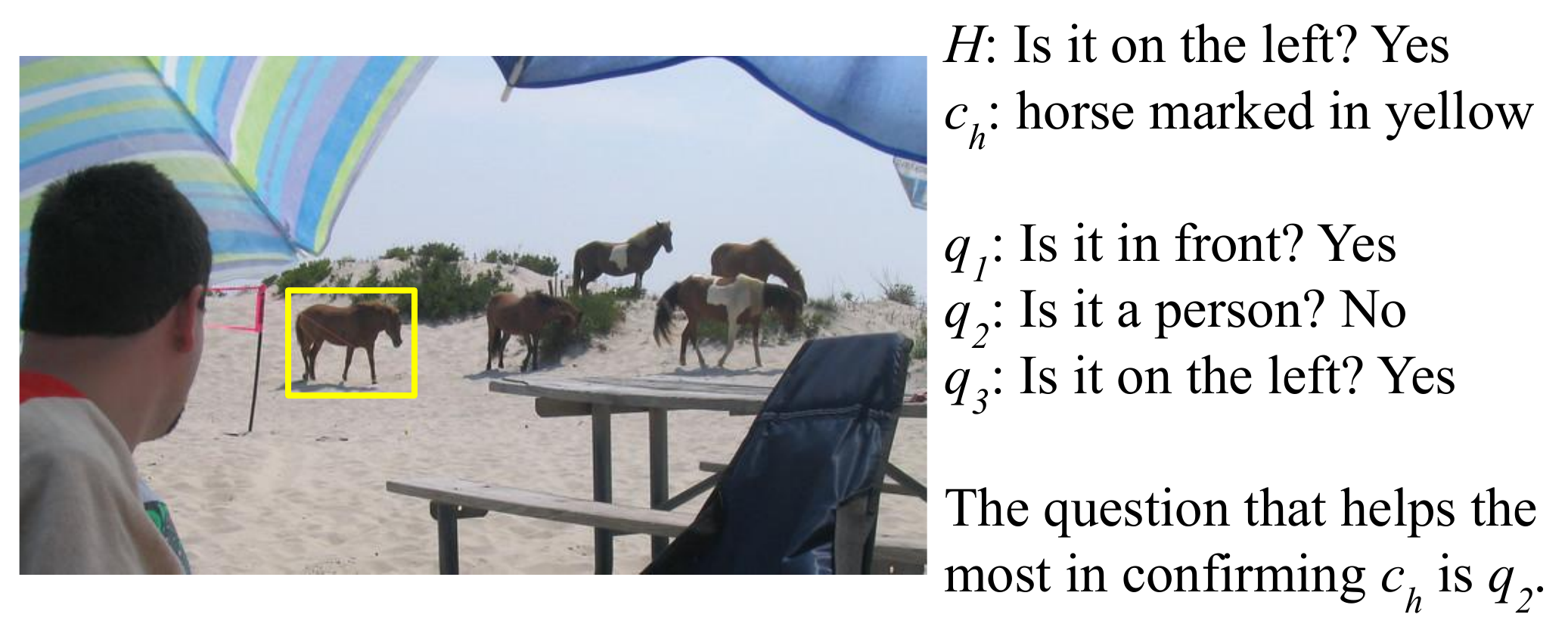}
	\caption{Given an image and the dialogue history $H$, \textit{Confirm-it} assigns the highest probability to $c_h$ (marked in yellow). Beam search generates three questions for the follow-up turn (ordered by their probability): thanks to its internal Oracle, the model anwers each of these questions by taking $c_h$ as the target. \textit{Confirm-it} selects $q_2$ (which receives a negative answer according to the internal Oracle) as the question that helps the most in confirming $c_h$.
	}
	\label{fig:question_with_no}
\end{figure}

\begin{figure*}[b]\centering
	\includegraphics[width=1\linewidth]{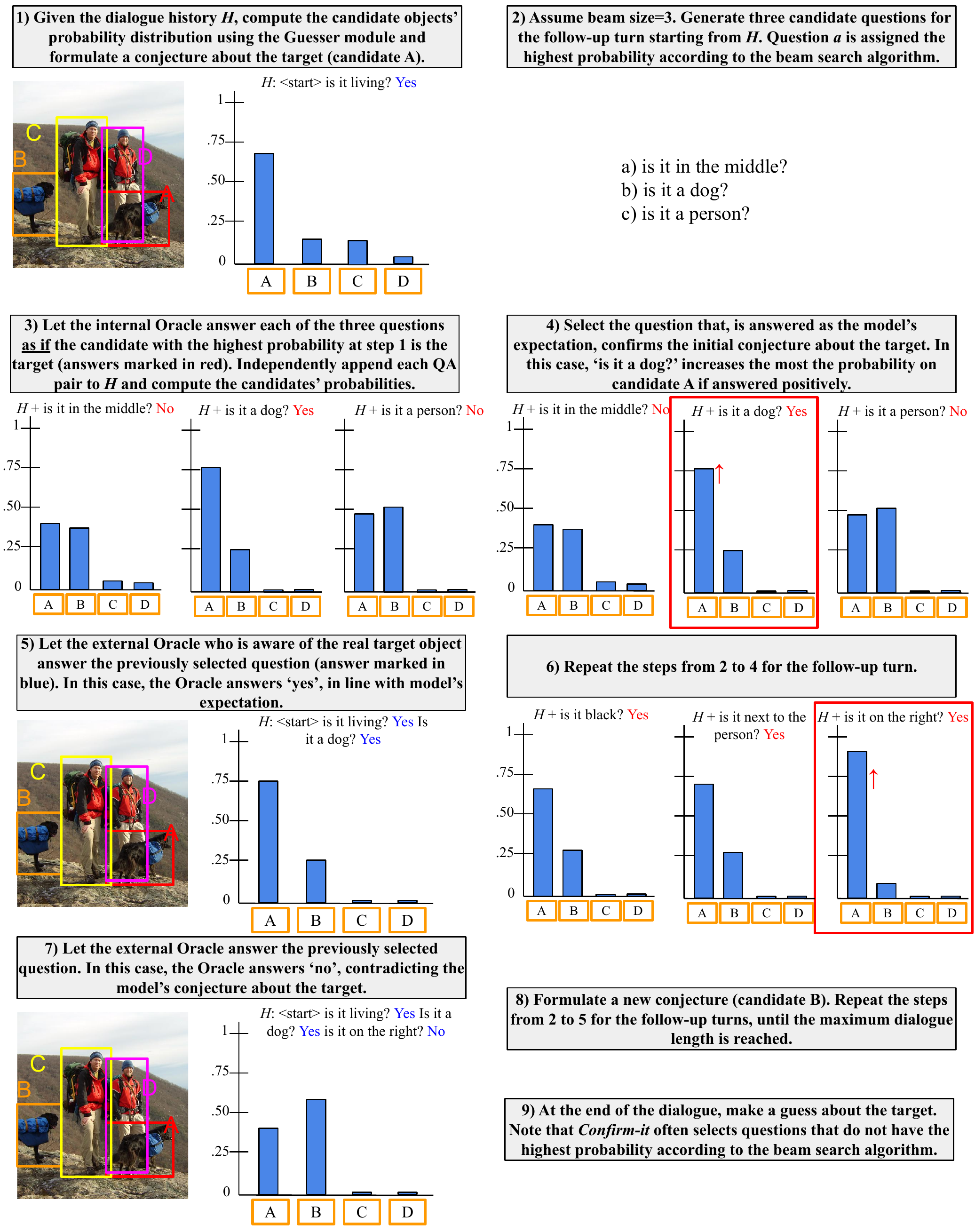}
	\caption{Step-by-step illustration of how \textit{Confirm-it} works.}
	\label{fig:confirm_it_step}
\end{figure*}

Section 4 of the paper describes the \textit{Confirm-it} model and Figure \ref{fig:gdse_v3} shows its architecture. In section 5 (Qualitative analysis of the strategy), we highlight that \textit{Confirm-it} does not select only questions for which it expects a positive answer, as shown in Figure \ref{fig:question_with_no}. In this case, given the dialogue history $H$, the model's hypothesis $c_h$ (the candidate that receives the highest probability according to the Guesser module), and a set of questions $q_1,q_2,q_3$ ordered by their probability according to beam search, the question that helps the most (answered by the internal Oracle taking $c_h$ as the target) is $q_2$.
Figure \ref{fig:confirm_it_step} illustrates a step-by-step example of how \textit{Confirm-it} works.

\paragraph{GuessWhat?!\@ Dataset Details.} The GuessWhat?!\@ dataset contains 155K English dialogues about approximately 66K different images from MSCOCO. The answers' distribution is: 52.2\% No, 45.6\% Yes, and 2.2\% N/A. The training set contains 108K dialogues and the validation and test sets 23K each. Dialogues contain on average 5.2 question-answer pairs. The vocabulary is built up of those words appearing at least 3 times in the training set, resulting in around 4900 words. Each game has at least 3 and at most 20 candidate objects. We train the model using human dialogues, selecting only the games on which humans have succeed in finding the target object in at most 10 turns (total number of dialogues used after this pre-processing step: 90K in training and around 18K both in validation and testing). The dataset of human dialogues is available at \hyperlink{https://guesswhat.ai/download}{https://guesswhat.ai/download}.

\paragraph{\textit{Confirm-it} Details.} Regarding the \textit{Confirm-it} model, we used a beam size of 3 in the paper and let the model generate dialogues of 5 turns. We also tried different values for the beam size, obtaining similar results. For the model hyperparameters and training procedure, we follow \citet{shekhar-etal-2019-beyond}. We trained and tested \textit{Confirm-it} on an NVIDIA TITAN V GPU (12 GB). We used Pytorch 1.0.1 (\hyperlink{https://pytorch.org/}{https://pytorch.org/}). \textit{Confirm-it} has 21411226 parameters. The average runtime is 15 minutes per epoch during training and 8 minutes during inference. The maximum number of training epochs is 100. We select the best model by looking at the performance on the GuessWhat?!\@ validation set. The validation accuracy of \textit{Confirm-it} is 51.49 (47.28 with beam search). For the CHAIR metric \cite{rohrbach-etal-2018-object}, we used the code in: \hyperlink{https://github.com/LisaAnne/Hallucination}{https://github.com/LisaAnne/Hallucination}.

\paragraph{Human Annotation Evaluation.} 
Figure \ref{fig:annotation_human} shows the annotation schema used by human participants in our study, as described in Section 5. The participants in this study are English proficient volunteers within our organization. Each participant is instructed on the guessing task by playing some trial games. The participant is admitted to the annotation only if he/she shows a clear understanding of the task. Given an image, a dialogue and a set of candidate objects with colour-matching boxes, participants express their guess by typing the number corresponding to the box of the selected candidate. Dialogues generated by human annotators from the GuessWhat?!\@ test set, by Confirm-it and by beam search without re-ranking were randomly presented.

\end{document}

%% file: introduction.tex
\section{Introduction}
\label{sec:introduction}


Quite important progress has been made on multimodal conversational
systems thanks to the introduction of the Encoder-Decoder
framework~\cite{suts:seq14}. The success of these systems can be
measured by evaluating them on 
task-oriented referential games. Despite the high task-success achieved
and the apparent linguistic
well-formedness of the single questions, the quality of the generated
dialogues, according to surface-level features, have been shown to be
poor; this holds for systems based on both greedy and beam search
(e.g.~\citet{shekhar-etal-2019-beyond, zarriess-schlangen-2018-decoding, murahari-etal-2019-improving}).
\citet{test:inter2021} found that
when taking these surface-level features as a proxy of linguistic quality, the
latter does not correlate with task
success, and the authors point to the importance of studying deeper features
of dialogue structures. We aim to develop a multimodal model able to
generate dialogues that resemble human dialogue strategies.  

Cognitive studies show that humans do not always act as ``rational''
agents. When referring to objects, they tend to be overspecific and prefer properties irrespectively of their utility for
identifying the referent~\cite{gatt2013bayesian};
when searching for information or when learning a language, they tend to follow
confirmation-driven strategies.  Modelling such behaviour in language learning,
\citet{medina2011words} and \citet{trueswell2013propose} propose 
a procedure in which a single hypothesized
word-referent pair is maintained across learning instances, and it is
abandoned only if the subsequent instance fails to confirm the
pairing. Inspired by
these theories, we propose a model, \textit{Confirm-it}, which
generates questions driven by the agent's confirmation bias.


\begin{figure}[t]\centering
	\includegraphics[width=1\linewidth]{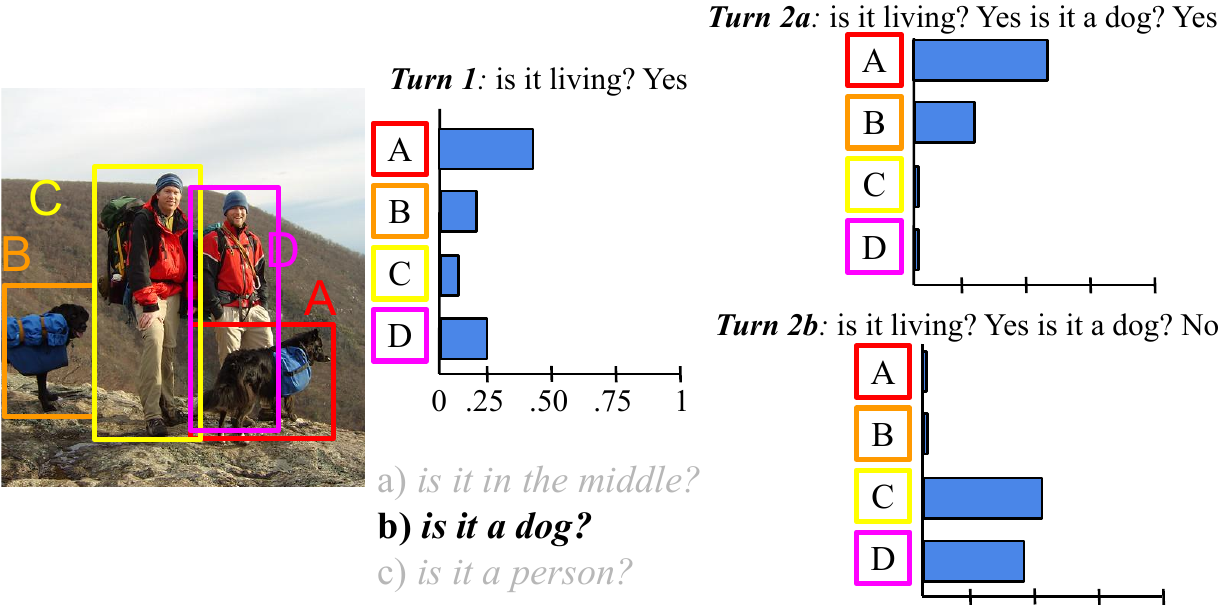}
	\caption{
		At turn 2, among the questions proposed by the beam
                search, \textit{Confirm-it}  chooses $b$  since it is the
                most suitable one to confirm the current conjecture.
		}
	\label{fig:intro}
\end{figure}

Take the example of a
referential guessing game in which an agent has to ask questions to
guess an object in a given image. \textit{Confirm-it} will ask questions
that reinforce its beliefs about which is the target object, till
proven otherwise. For instance, in Figure~\ref{fig:intro}, after
learning that the target is a living entity (turn 1), the agent conjectures
the target is the dog on the right of the picture (though in
principle, it could have been any of the candidates). Hence, the decoder generates the question that would let it confirm such
belief, ``is it a dog?''. If its expectations are not met (viz., it
receives a negative answer to such question - turn 2b), it moves its attention
to another candidate object.  
We do not claim that
our choice represents the optimal strategy to play the game,
but we believe that it makes the generated dialogue more human-like. 

To evaluate this strategy, we take as a test-bed GuessWhat?!\@
\cite{guesswhat_game}, a two-player game between a Questioner that has to guess the target, and an Oracle (called ``external Oracle'' in the following) who is aware of the target.  
The widely used architecture of the Questioner, GDSE, jointly trains a Question Generator (QGen) and a Guesser~\cite{shekhar-etal-2019-beyond}. We augment this architecture with a module that simulates an internal Oracle. Being an ``internal'' Oracle, at test time this agent does not know what the target object is: while at training time it learns to answer questions by receiving the gold standard datapoint (the question, the actual target, and the human answer), at test time it assumes the target is the candidate object to which the Guesser assigns the highest probability.  Hence, the three modules of the Questioner straightforwardly cooperate one another. 
The internal Oracle guides the QGen to ask questions that reinforce the Guesser's beliefs. Concretely, at training time, through Supervised Learning (SL) the QGen learns to ask human-like questions turn-by-turn, the internal Oracle to answer them, and the Guesser to guess the target object once the dialogue ends. At test time, we implement a beam search re-ranking
algorithm that simulates the single-conjecture learning strategy used
by humans: among the questions the QGen 
generates via beam search, the algorithm promotes the questions whose answer (obtained via the internal Oracle that receives the candidate with the highest probability as the target) increases the most the model's confidence in its hypothesis about the target.




We run both quantitative and qualitative analyses, and evaluate the
effectiveness of the dialogue strategy by asking human annotators to guess the
target object given the dialogues generated by \textit{Confirm-it}. We compare
results giving the dialogue generated by \textit{Confirm-it}
when using the re-ranking algorithm and when generating the question
proposed by the plain beam search. We show
that the task accuracy of both
the conversational agent and human subjects increases when receiving the dialogues generated by the \textit{Confirm-it} re-ranking algorithm.





%% file: related.tex
\section{Related Work}
\label{sec:related}


For open-ended language generation, \citet{holt:thecu20} claim that
decoding strategies that optimize for output with high probability
(like beam search) lead to highly deteriorated texts, since the
highest scores are often assigned to generic, incoherent, and
repetitive sequences. Several works propose reranking strategies on the
set of hypotheses produced by the beam search following different
criteria~\cite{dusek-jurcicek-2016-sequence,blain2017exploring,agarwal-etal-2018-char2char,BorgeaudE20,HargreavesVE21} to improve both the performance on a given task and the quality of the output. In this work, we present a cognitively-inspired reranking technique for a visual dialogue questioner agent.

In visual dialogue systems, the quality of the output has been
improved mainly by aiming at reducing repetitions in the output. This goal has been achieved through Reinforcement Learning by adding auxiliary objective
functions~\cite{murahari-etal-2019-improving}, intermediate rewards
\cite{zang18}, regularized information gain techniques
\cite{shukla-etal-2019-ask}, or intermediate probabilities with an
attention mechanism \cite{pang_gw}.
Different from these works, we do not use the
Reinforcement Learning paradigm and, instead of focusing on improving
surface-level features, we indirectly operate on the
dialogue structure.


\citet{ruggeri2015children} studied the way children and young adults search for information while asking yes-no questions given a set of candidate hypotheses. The authors found that when prior knowledge favours some hypotheses over others, participants asked more hypothesis-scanning questions (i.e., questions that are tentative solutions, with a specific hypothesis that is directly tested). This is in line with the observation in~\citet{baron:think10} that humans phrase questions to receive an affirmative
answer that supports their theory, and with the broader finding in ~\citet{wason60} that they tend to select the information that is in accord
with their prior beliefs. Inspired by
these studies, we propose a new dialogue strategy for playing
referential guessing games by exploiting the probabilities assigned by the Guesser module to different candidate objects.

\cut{
This is line with the more general observation on humans being driven by confirmation (or verification) biases. For instance, when searching
for information, they phrase questions to receive an affirmative
answer that supports their theory~\cite{baron:think10}. 

In other words, they
tend to select the information that supports their view, that is in accord
with their prior beliefs~\cite{wason60}.}

\cut{This widely used confirmation-driven approach has been found also in
cognitive studies of language learning. 
These studies claim that humans follow the single-conjecture learning
procedure when acquiring a new word meaning. In other words, learners hypothesize a single word-referent pairing based on
conceptual and referential biases and abandon such hypothesis only if
the subsequent instance fails to confirm such pairing (e.g. \citealt{medina2011words,trueswell2013propose}).}

%% file: task_data.tex
\section{Task and Dataset}
\label{sec:task_data}

GuessWhat?!\@ \citep{guesswhat_game} is an asymmetric game involving two human
participants who see a real-world image from MSCOCO
\cite{lin:2014}. One of the participants (the Oracle) is secretly
assigned a target object in the image, while the other participant
(the Questioner) has to guess it by asking binary (Yes/No) questions
to the Oracle. The GuessWhat?!\@ dataset consists of more than 150k
human-human English dialogues containing on average 5.3 questions per dialogue.

%% file: model_experiments.tex
\section{Model and Re-ranking Strategy}
\label{sec:model_experiments}


Our model, \textit{Confirm-it}, builds on
GDSE~\cite{shekhar-etal-2019-beyond}. In the latter, the hidden state
representation produced by a multimodal encoder is used to jointly
train the question generator (QGen) and the Guesser module. The image is encoded with a ResNet-152
network~\cite{he2016:resnet} and the dialogue history is encoded via
an LSTM network. QGen uses greedy search to generate questions. To
this multi-tasking setting, \textit{Confirm-it} adds an internal
Oracle trained to answer the question at each turn.  Moreover, it
relies on beam search and, at inference time, it goes through a
re-ranking phase which simulates the single-conjecture learning
strategy. The model architecture is provided in the Supplementary
Material (SM) and the algorithm is spelled out below.

\begin{algorithm}
	\caption{The \textit{Confirm-it} algorithm}\label{alg:confirm}
	\begin{algorithmic}[1]
		\REQUIRE History $H$, Beam size $B$, Max turns $T$, Image $I$, Distractors $D_{1:N-1}$, target $o_t$,
		\REQUIRE Candidates $C_{1:N} \gets D_{1:N-1}+o_t$
		\REQUIRE Internal Oracle $IO$
		\REQUIRE Target-aware external $Oracle$
		
		\FOR{$turn = 1:T$} 
		\STATE $\overset{\wedge}p(c_{k_{1:N}}) \gets Guesser(H, I, C_{1:N})$
		\STATE $c_{h} \gets argmax(\overset{\wedge}p(c_{k_{1:N}}))$ 
		
		\STATE $q_{1:B} \gets QGen(H, I)$
		\STATE $a_{1:B}^* \gets IO(H+q_{1:B}, c_{h})$
		\STATE $H^*_{1:B} = H + (q_{1:B}, a_{1:B}^*)$
		\STATE $p^* \gets Guesser(H^*_{1:B}, I, C_{1:N})$
		\STATE $Q \gets q_{1:B}[argmax(p^*(c_{h}))]$	
		\STATE $Oracle$ provides an answer $A$ to $Q$
		\STATE $H \gets H+Q+A$
		\ENDFOR
		
	\end{algorithmic}

\end{algorithm}


\cut{\citet{shekhar-etal-2019-beyond} propose the Visually-Grounded Dialogue State Encoder model (GDSE). This model uses a visually grounded dialogue state that takes the visual features of the input image and each question-answer pair in the dialogue history to create a shared representation used both for generating a follow-up question (QGen module) and guessing the target object (Guesser module) in a multi-task learning scenario. More specifically, the visual features are extracted with a ResNet-152 network \cite{he2016:resnet} and the dialogue history is encoded with an LSTM network. In this work, we augment the GDSE architecture with the Oracle model so that all the three tasks involved in GuessWhat?!\@(asking questions, guessing the target, answering questions) are performed by the same model (GDSE-v3). We adopt the \textit{modulo-n} training proposed in \citet{shekhar-etal-2019-beyond} to address the issue of dealing with tasks with different levels of difficulty. We use $n=7$ and we train the Oracle module with the same frequency as the Guesser module. In the Supplementary Material (SM) we report the model's architecture.}


Algorithm \ref{alg:confirm} describes the beam search re-ranking
algorithm used by \textit{Confirm-it} to promote the generation of an
effective dialogue strategy. Given an image, a set of candidate objects, a
target object $o_t$, and a beam size of $B$, at each dialogue turn the
model predicts a probability distribution over the set of candidate
objects given the current dialogue history. The candidate that
receives the highest probability is considered the model's
\textit{hypothesis} $c_{h}$. The QGen outputs $B$
questions, ordered by their probability. Each of these questions is
answered by the model's internal Oracle that receives $c_{h}$ as the
target object. Among these $B$ questions, \textit{Confirm-it} selects
the question $Q$ that, paired with the answer provided by the internal
Oracle, increases the most the model's confidence over $c_{h}$,
measured as the probability assigned by the Guesser. The external
Oracle (who is aware of the real target object $o_t$) answers $Q$, and
this new question-answer pair is appended to the dialogue history.
In SM we provide a step-by-step example of how \textit{Confirm-it} works.

None of the features of our case-study are crucial for the method to be applied to other tasks, e.g. it does need the questions to be polar, it does not need the questions to be visually grounded, it does not need the dialogue to be asymmetrical.

\paragraph{Implementation details\footnote{Our code is available at: \url{https://github.com/albertotestoni/confirm_it}}}
For the multi-task training, we adopt the \textit{modulo-n} training proposed in
\citet{shekhar-etal-2019-beyond}, i.e. we train the Oracle and guesser
modules every $n$ (=7) epochs of the QGen.  
At inference time, we use a beam size of 3 and let the model generate dialogues of 5 turns.

%% file: results.tex
\section{Results}
\label{sec:results}

\begin{table}[]
	\centering
	\begin{tabular}{c|c}
		\hline
		\textbf{Decoding Strategy} & \textbf{Accuracy 5Q} \\ \hline
	\textit{Confirm-it} & 51.38 \\ \hline
		Beam Search & 47.03 \\ \hline
Random Re-Rank & 46.29 \\ \hline		
Greedy Search & 46.58 \\ \hline
		
	\end{tabular}
	\caption{Accuracy results of \textit{Confirm-it} on the GuessWhat?!\@ test set when generating 5 questions per dialogue following different decoding strategies. \textit{Confirm-it} refers to the accuracy achieved by our re-ranking algorithm compare to a random re-ranking of candidate questions, plain beam search, and greedy search.}
	\label{tab:accuracy}
\end{table}


\begin{table*}[]
	\centering
	\small
	\begin{tabular}{|c|c|c|c|c|c|c|}
		\hline
		& \begin{tabular}[c]{@{}c@{}}Human\\ Accuracy $\uparrow$ \end{tabular} & \begin{tabular}[c]{@{}c@{}}\% games with\\ repetitions $\downarrow$ \end{tabular}  & CHAIR-s $\downarrow$ & \begin{tabular}[c]{@{}c@{}}\% Yes\\ Last Turn$\uparrow$ \end{tabular} & \begin{tabular}[c]{@{}c@{}} \% novel $(q_{t-1},q_t)$\\ per dialogue $\uparrow$\end{tabular}  \\ \hline
		Beam Search& 70.8 & 38.50 & 31.07 & 71.87 & 36.75 \\ \hline
		\textit{Confirm-it}& 77.0 & 30.36 & 28.46 & 76.68 & 48.25 \\ \hline
Human Dialogues & 96.0 & 0.01& 7.45 & 86.64 &-\\\hline
	\end{tabular}
	\caption{ 
	\textit{Human Accuracy} refers to the task accuracy achieved by human annotators when receiving dialogues generated by the plain Beam Search, \textit{Confirm-it} re-ranking, or the original dialogues produced by human players from the GuessWhat?! test set. The other columns report relevant statistics of the dialogues: percentage of games with at least one repeated question verbatim, hallucination rate (CHAIR-s), percentage of positive answers in the final turn (\% yes Last Turn), and percentage of consecutive questions not seen at training time (lexical overlap, \% novel $q_{t-1},q_t$ per dialogue).
	}
	\label{tab:analysis}
\end{table*}

We study to what extent the re-ranking phase lets the model generate
more effective and more natural dialogues. To this end, we evaluate the
\textit{Confirm-it} task-accuracy with and without the ranking
phase\footnote{Remember that \textit{Confirm-it} is based on the GDSE architecture. For comparison, the accuracy reached by GDSE is 45.55\% with greedy search and 46.40\% using beam search decoding.} and report qualitative analyses of the
dialogues. 

\paragraph{Task-accuracy} Table~\ref{tab:accuracy} shows the task accuracy
results of our model in the GuessWhat?!\@ game.  When the model undergoes the re-ranking phase,
\textit{Confirm-it} accuracy has an increase of +4.35\% with respect to
what it achieves when it outputs the question selected by the plain beam search,
and an increase of +4.8\% against greedy search.  Note that, instead, randomly
re-ranking the set of questions  lowers the performance.  This
result shows that confirmation-driven strategies help generate more
effective dialogues. Interestingly, our re-ranking method does not require
additional training compared to the SL paradigm.

\begin{figure}[t]\centering
	\includegraphics[width=0.93\linewidth]{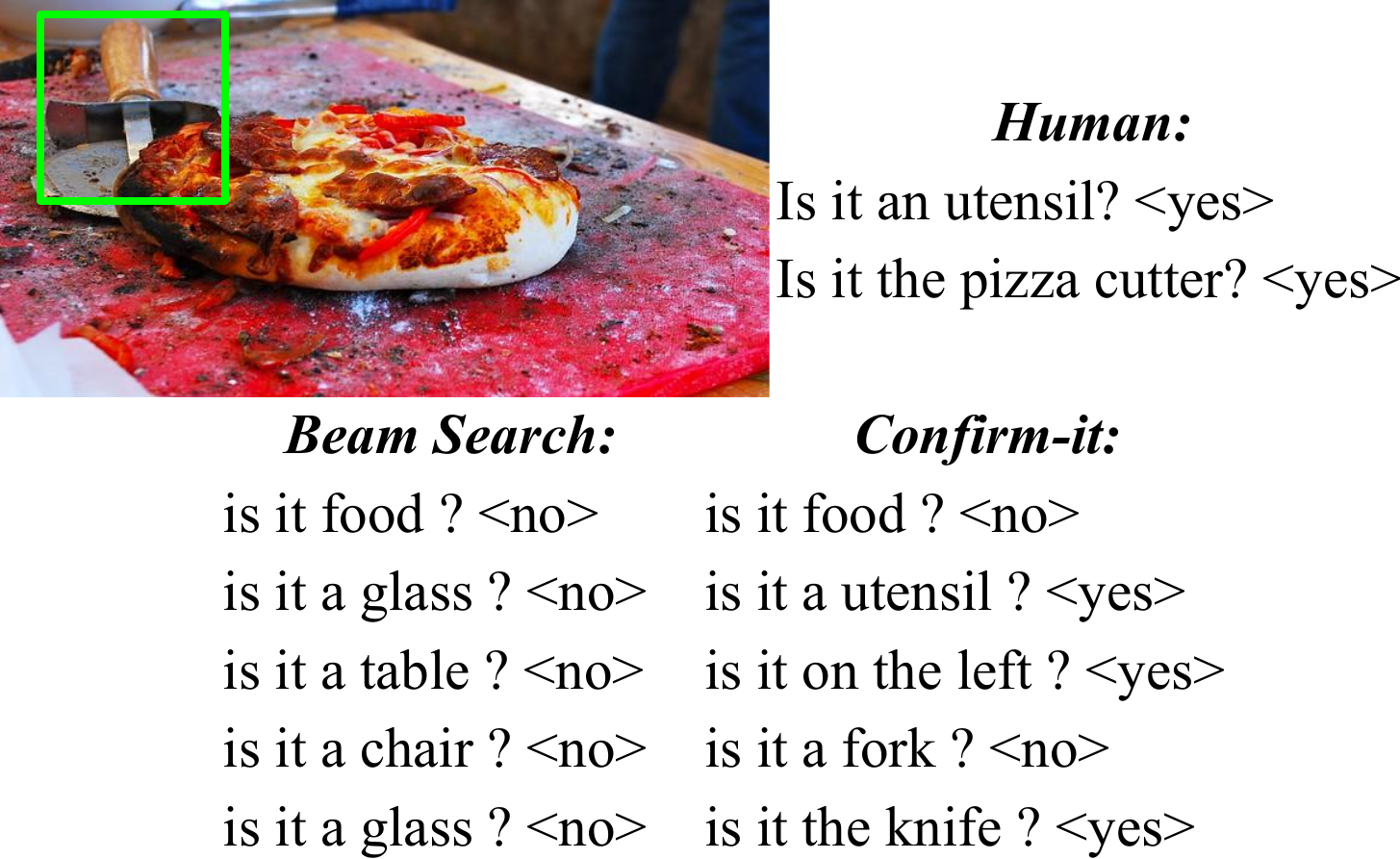}
	\caption{Through re-ranking, dialogues become more effective
          and more natural. The target object is highlighted in green.}
	\label{fig:comparison}
\end{figure}

\paragraph{More Effective Dialogues} To verify whether the improvement
of \textit{Confirm-it} is really due to the generation of
more effective dialogues to solve the guessing task,
we asked human subjects to guess the target given a dialogue. We
sampled 500 games from the GuessWhat?!\@ test set containing less than
7 candidate objects. Each participant played 150 games equally divided
among dialogues
generated by the model with the plain beam search, with our re-ranking strategy, and by the GuessWhat?!\@ 
human players (taken from the original test set). We made
sure no participant played the same game more than once.
In total, 10 English proficient volunteers within our
organization joined the experiment. As we can see from Table
\ref{tab:analysis}, human annotators reach an accuracy of 70.8\% in identifying
the target object when receiving dialogues generated by beam search
and 77\% with \textit{Confirm-it}, suggesting that the re-ranking
phase let the model generate more effective dialogues. The accuracy that the annotators achieve when playing the game with dialogues extracted from the original GuessWhat?!\@ test set (and thus generated by human players) is much higher (96\%).



\cut{In total, 10 English proficient volunteers within our organization joined the experiment. They were instructed on the GuessWhat?!\@ task and each participant played on average 50 games with dialogues generated by beam search and 50 by \textit{Confirm-it}. }

\cut{\at{To  be removed? -->} Moreover, we asked participants to compare two dialogues (one generated with beam search and one with our algorithm) given an image and a target object and to select the dialogue that sounds more natural to identify the target. We found that in the majority of the cases, humans preferred dialogues generated by \textit{Confirm-it} (54.2\% vs. 45.8\%). In SM, we report an example of the annotation carried out by human subjects. }

\cut{\at{ Forse sono io che devo ancora metabolizzare, ma non mi torna la descrizione dei risultati se qui non scriviamo qualcosa tipo: In the following, we use ``Beam Search'' to present the results obtained with a beam search decoding without re-ranking. We use instead ``\textit{Confirm-it}'' for the results obtained with the re-ranking algorithm described above.}}


\paragraph{More Natural Dialogues}
Figure~\ref{fig:comparison} reports a sample game that illustrates the
difference between a dialogue generated by human players, one generated by the plain beam search, and one by our re-ranking algorithm. The dialogue generated by beam search contains a repetition
(``is it a glass?''), it asks about entities not present in the
image (``chair'' and ``glasses'') and it ends with a non-conclusive negatively
answered question. These features contribute to making the dialogues
sound unnatural.
We check whether the re-ranking phase
helps our model to get closer to human dialogues with respect to these
features.  To this end, we compute the percentage of games with
repeated questions and with the last turn containing a positively
answered question.  Moreover, we employ CHAIR-s to measure the
percentage of hallucinated entities in a
sequence, originally proposed in ~\citet{rohrbach-etal-2018-object} for image captioning and recently applied also to GuessWhat?!\@ \cite{testoni-bernardi-2021-ive}. CHAIR-S is defined as the number of dialogues with at last one hallucinated entity divided by the total number of dialogues. As we can see from Table~\ref{tab:analysis}, dialogues generated by \textit{Confirm-it} contain
fewer games with at least one repeated question compared to the beam
search setting (-8.14\%), fewer games with hallucinated entities (-2.61\%)\footnote{\citet{rohrbach-etal-2018-object} propose another variant of the CHAIR metric called CHAIR-i (per-instance), defined as the number of hallucinated objects in a sequence divided by the total number of objects mentioned. CHAIR-i results: 18.28 (Beam Search), 15.02
  (\textit{Confirm-it}), 4.11 (Human Dialogues).}, and more games with the last
turn containing a positively answered question (71.87\% vs. 76.68\%). 
The reduced number of hallucinations is a direct consequence of the \textit{Confirm-it} strategy: following up on a single object through the dialogue, the model is less likely to engage in spurious exchanges on irrelevant objects. Though this strategy continuously looks for confirmations, it is worth noting that it does not increase the number of repetitions, which instead are significantly reduced. This is an interesting property emerging from the interplay between the internal Oracle and the re-ranking strategy, which suggests that asking the very same question more than once in a dialogue does not increase the model's confidence in its hypothesis.

\paragraph{Qualitative Analysis of the Strategy}
We also evaluated the strategy followed by \textit{Confirm-it} by
looking at the model's decisions throughout the
dialogue. Interestingly, the model does not select only questions for
which it expects a positive answer, though they are the majority
(67\%).  See the SM, for a game in which the
re-ranking promoted a question answered negatively by the internal Oracle. Moreover,
though the model looks for confirmations, it properly updates its
beliefs when disconfirmed: when the model receives from its
interlocutor an answer different from the one it expects (based on its
internal Oracle), in 70\% of the cases the Guesser changes the probabilities
over the candidates accordingly, i.e., it assigns the highest probability to a new candidate object. Finally, the use of a human-like strategy does not imply having learned to simply mimic human dialogues from the training set:
the re-ranking shows an absolute increase of +12\% in the number of pairs of consecutive questions not
seen during training (see Table~\ref{tab:analysis}).


%% file: conclusion.tex
\section{Discussion and Conclusion}
\label{sec:conclusion}

In this paper, we propose \textit{Confirm-it}, a multimodal
conversational model based on a decoding strategy inspired by
cognitive studies of human behavior.  We show that, through the
proposed beam search re-reanking algorithm, our model generates dialogues that
are more effective (based on task-accuracy) and more natural (based on
the dialogues features discussed above). We believe further improvement could be obtained by
increasing the performance of every single module. Moreover, the
structure of the generated dialogues remains to be analysed, and we agree
with~\citet{csleval2020} that a proper evaluation should involve humans. In future work, our method can be easily extended to other task-oriented dialogue tasks which involve a conversational agent as far as it has a module that generates questions and a module that performs a classification task. Depending on the task at hand, different ways to take intermediate probabilities into account can be designed, but the core idea of the method would not change.